\documentclass[final,3p,times,twocolumn]{elsarticle}

\usepackage{amssymb}
\usepackage{amsmath}
\usepackage{booktabs}
\usepackage[colorlinks=true,citecolor=blue,linkcolor=blue,urlcolor=blue]{hyperref}

\graphicspath{{figures/}}

\journal{Journal Name}

\begin{document}

\begin{frontmatter}

\title{Automated Nucleus Segmentation in Histology Images: An Attention U-Net Approach with Comprehensive Preprocessing and Inference Pipelines}

\author{Akshat G}
\author{Divyansh Gupta}
\author{Tusar Kanti Mishra\corref{cor1}}
\ead{tusar.mishra@manipal.edu}
\cortext[cor1]{Corresponding author}

\affiliation{organization={Manipal Institute of Technology Bengaluru, Manipal Academy of Higher Education},
            city={Manipal},
            country={India}}

\begin{abstract}
\textbf{Objectives}\\
In the area of digital pathology, nucleus segmentation is an essential step for quantitative analysis; however, high nuclear density and staining heterogeneity in histology images pose a great challenge.

\medskip
\noindent\textbf{Methods}\\
This study implements a lightweight Attention U-Net model on the MoNuSeg 2018 dataset to overcome these challenges. The proposed pipeline ties the comprehensive image and mask preprocessing-including stain deconvolution and multi-pass mask smoothing with a high-fidelity overlapping patch-stitching strategy at the inference stage on high resolution images.

\medskip
\noindent\textbf{Results}\\
The model achieved Intersection over Union (IoU) score of 0.8745 on full resolution, unseen test images. This result confirms the effectiveness of using attention mechanisms to enhance feature localization and verifies the robust stitching methodology for generating seamless and accurate segmentation maps.

\medskip
\noindent\textbf{Conclusion}\\
Attention U-Net is a very effective and highly competitive method for automated, scalable nucleus delineation in challenging clinical context.
\end{abstract}


\end{frontmatter}


\section{Introduction}
\label{sec:intro}

Accurate segmentation of cell nuclei is a fundamental task in digital pathology, offering reliable quantitative analysis for clinical diagnosis and biological studies. The automatic delineation of individual nuclei enables the extraction of important morphometric features, including form, size, and texture, which are critical indicators of tumor grading and patient prognosis. However, this task poses several challenges in computer vision. Histological images often exhibit high variability in the appearance of cell nuclei, non-uniform staining, and high-density cell clusters with frequent overlaps, which require very sophisticated and generally applicable algorithms to prevent various types of segmentation errors.

While early efforts relied on traditional methods of computer vision, especially U-Net-based architectures, deep learning techniques marked a leap in robustness and performance. Building upon this success, this paper discusses the implementation and evaluation of a lightweight Attention U-Net model that uses attention gates to help a model concentrate on the most important characteristics for the segmentation task. We will be using the 2018 Multi-Organ Nucleus Segmentation (MoNuSeg) Challenge dataset to validate this approach.

This work presents an end-to-end pipeline for applying the Attention U-Net model to the MoNuSeg dataset, covering data preprocessing, model training, full-image inference, and a comprehensive evaluation of its performance. The subsequent sections outline in detail the image and mask preprocessing to enhance them, the proposed model architecture along with the training strategy, and also a thorough experimental analysis including both quantitative metrics and qualitative visual results.

\subsection{Literature Review}
\label{subsec:litreview}

In computational pathology, segmentation of cell nuclei using automated algorithms from histopathology images represents an important problem and one that has traditionally posed considerable challenges. Traditional non-deep learning-based approaches include intensity thresholding, watershed algorithms, and active contours. These were significantly limited, however, by their reliance on manually designed features and pre-specified criteria, thus could not adapt to the variability in nuclear appearance within different tissues, staining protocols, and disease conditions. Consequently, they usually failed under variable staining, overlapping, or clumped nuclei with noisy backgrounds, which resulted in poor generalization and poor segmentation, especially in crowded or chromatin-sparse regions \cite{b1,b2}.

This paradigm shifted with the introduction of deep learning, in particular Fully Convolutional Networks, which began to overcome these challenges by directly learning hierarchical features in a data-driven fashion from large annotated datasets. This allowed models to adapt far more effectively to complex and variable nuclear morphologies \cite{b1,b2}. The U-Net architecture was among the first and rapidly established the gold standard in biomedical image segmentation. U-Net introduced a symmetric encoder-decoder structure combined with a novel mechanism of skip connections. The idea behind it is straightforward: while the encoder path progressively down-samples the input to capture abstract contextual data, the decoder path works to restore spatial resolution. Importantly, the skip connections transfer high-resolution feature maps from the encoder directly to the corresponding layers in the decoder. Such a fusion of detailed spatial information with high-level contextual features proved exceptionally effective for achieving precise localization of nuclei boundaries, where preserving fine-grained precision is crucial \cite{b3,b4,b5}.

While powerful, the skip connections in a standard U-Net can accidentally propagate redundant or irrelevant low-level feature information in complex situations. In order to overcome this, attention mechanisms have recently been incorporated into models such as the Attention U-Net. These models include attention gates in the skip pathways that learn to selectively emphasize salient features that contribute toward segmentation and suppress background noise and less relevant activations. The attention gate looks at both encoder features and decoder context to compute an attention map that is used in reweighting the feature maps before fusion. This ensures that only the most informative features are propagated, hence enhancing the precision and robustness of the segmentation. \cite{b6,b7,b8}.

Besides attention mechanisms, other influential U-Net variants have been developed focusing on further improving feature representation. For example, U-Net++ uses nested and dense skip pathways to build a set of intermediate convolutional layers that progressively decrease the semantic gap between encoder and decoder, allowing for richer multi-scale feature aggregation \cite{b9,b4}. On the other hand, models like TransUNet implement a transformer-based encoder and a Convolutional Neural Network (CNN) based decoder. Such a hybrid architecture combines the transformer's modeling abilities of long-range dependencies and global context with the CNN's ability to capture fine-grained local features, enabling a complete understanding of the image. This is particularly effective for tasks that require both precise localization and broad contextual awareness. \cite{b10,b11}.

The availability of high-quality data aided in the rapid development and validation of architectures mentioned prior, on publicly available datasets. One of such datasets is the Multi-Organ Nucleus Segmentation (MoNuSeg -2018) \cite{b12} Challenge dataset, which aimed at testing model generalizability in the field of multi tissue segmentation task \cite{b12}. MoNuSAC2020, CryoNuSeg and other prominent datasets of cryosectioned images. To derive rational conclusions across studies, the domain uses a set of standard evaluation benchmarks. Dice Coefficient \cite{b18} and Jaccard Index (IoU) \cite{b19} are the most common, measuring the degree of overlap between predicted and ground truth masks to access the overall accuracy of the predicted segmentation region. Together, the above mentioned datasets and metrics provide a robust platform to objectively assessing the performance of segmentation algorithms \cite{b2,b12,b13,b16}.

\section{Methods}
\label{sec:methods}

\subsection{Preprocessing}
\label{subsec:preprocessing}

A rigorous preprocessing pipeline was developed to regularize the input data and highlight features relevant in nucleus segmentation. In order to enhance data coherence and optimize model performance, several tailored transformations have been performed on both the raw histology images and their ground truth masks. These two pipe-lines were executed in parallel, with particular attention to maintaining a perfect preservation of spatial relationships between an image and its mask across all steps. This will be important in highlighting the cellular nuclei and preparing the data for the subsequent stage of patch generation.

\begin{figure}[!htbp]
\centering
\includegraphics[width=0.45\textwidth,height=0.42\textheight,keepaspectratio]{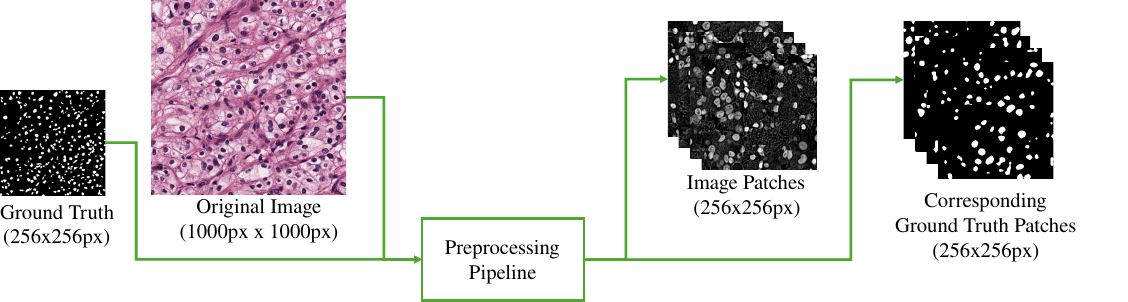}
\caption{High-level representation of the preprocessing pipeline.}
\label{fig1}
\end{figure}

\subsubsection{Image Preprocessing Pipeline}
\label{subsubsec:imgprep}

The raw .tif histology images underwent an enhancement process which comprised of color normalization to normalize their appearance and accentuate the nuclei. The color normalization process was centered around the isolation of the Hematoxylin channel by stain deconvolution, a process that transforms an image into a single channel, with emphasis on nuclear structures. Subsequently, a set of standard image enhancement techniques such as Median Filtering for noise reduction, Contrast Limited Adaptive Histogram Equalization (CLAHE) to enhance local contrast, and Unsharp Masking to highlight nuclear boundaries were employed. The result of such an automated pipeline is clean, high-contrast images optimized for the subsequent segmentation task.

\begin{figure}[!htbp]
\centering
\includegraphics[width=0.45\textwidth,height=0.42\textheight,keepaspectratio]{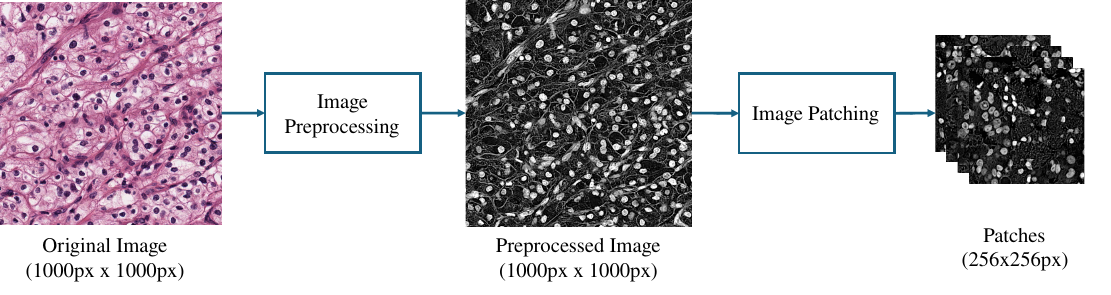}
\caption{Modular representation of image preprocessing pipeline.}
\label{fig2}
\end{figure}

\subsubsection{Mask Preprocessing Pipeline}
\label{subsubsec:maskprep}

For the masks, representing ground truth, a different preprocessing pipeline was needed to transform them into appropriate form and format for training. The key objectives were matching spatial dimensions to the full-size images and creating clean binary target masks with smooth contours. This was achieved by performing the following steps:

\begin{enumerate}
\item \textbf{Spatial Upsampling:} Original masks were upsampled from 256x256 to 1000x1000 pixels to allow proper 1:1 pixel correspondence with the pre-processed histology images; it was done using the cv2.INTER\_NEAREST interpolation method to keep the hard boundaries of the annotated nuclei without introducing any ambiguous, anti-aliased pixel values.
\item \textbf{Refinement and Binarization:} To get a smooth and correct binary ground truth, a multi-pass routine was performed. It consisted of the iterative application of a Gaussian blur on the edges of the upsampled masks to smoothen them, followed by a binary thresholding operation. This step effectively removed small artifacts while leaving the final masks with only two pixel values, 0 for the background and 1 for the nuclei, hence providing a smooth mask for the model to learn from.
\end{enumerate}

\begin{figure}[!htbp]
\centering
\includegraphics[width=0.45\textwidth,height=0.42\textheight,keepaspectratio]{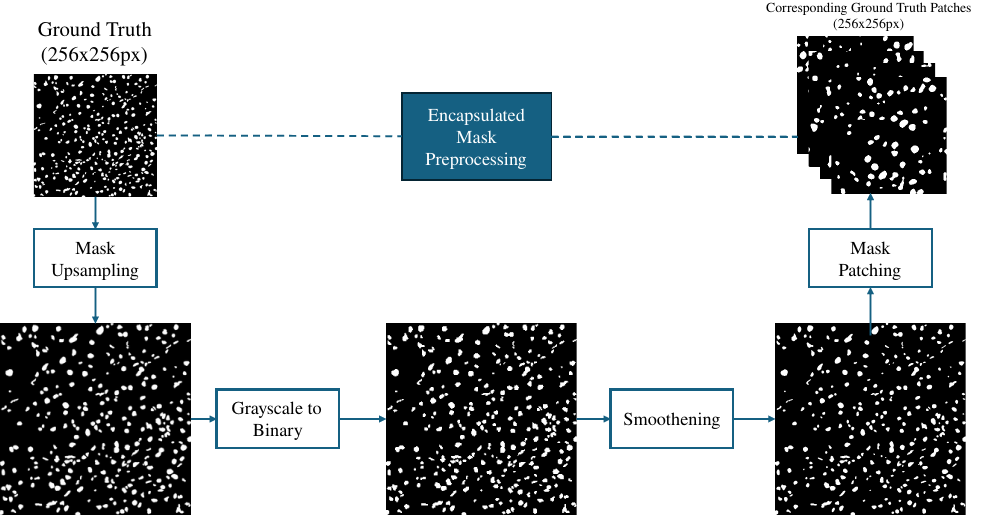}
\caption{Modular representation of mask preprocessing pipeline.}
\label{fig3}
\end{figure}

\subsubsection{Patch Generation and Data Splitting}
\label{subsubsec:patches}

Because of the high resolution of the images and typical GPU memory, the model was trained on smaller patches of 256x256 pixels rather than full images of size 1000x1000. This also serves as augmentation and increases the number of training samples. To ensure a fair evaluation, the original set of 30 full-size images was first partitioned. 24 images (80\%) were allocated for the training set, while the remaining 6 images (20\%) were held out as a test set for final evaluation. Subsequently, 100 patches were generated from each of the 24 training images by randomly sampling 256x256 regions. The corresponding mask patch was extracted from exactly the same coordinates so that the input image and ground truth are perfectly aligned. This resulted in a final training dataset of approximately 2,400 image-mask pairs.

\subsection{Proposed Methodology}
\label{subsec:methodology}

\subsubsection{Model Architecture}
\label{subsubsec:architecture}

The model used is a lightweight Attention U-Net \cite{b20} with approximately 2M parameters, which extends the original encoder-decoder U-Net architecture \cite{b21} with attention gates on its skip connections. The network processes a (256, 256, 1) input via a contracting path consisting of four blocks; each block executes two 3x3 convolutions, followed by max-pooling, to consecutively reduce spatial dimensions while increasing feature channels from 16 to 128. After a bottleneck layer with 256 channels, the resolution is restored through a symmetric expansive path. More importantly, before the skip connections in each decoding step are concatenated, an attention gate re-weights the features produced by the encoder. This enables the model to focus only on the most informative input for segmentation. Finally, the required probability map comes from a 1x1 convolution followed by sigmoid activation.

\begin{figure}[!htbp]
\centering
\includegraphics[width=0.5\textwidth,height=0.5\textheight,keepaspectratio]{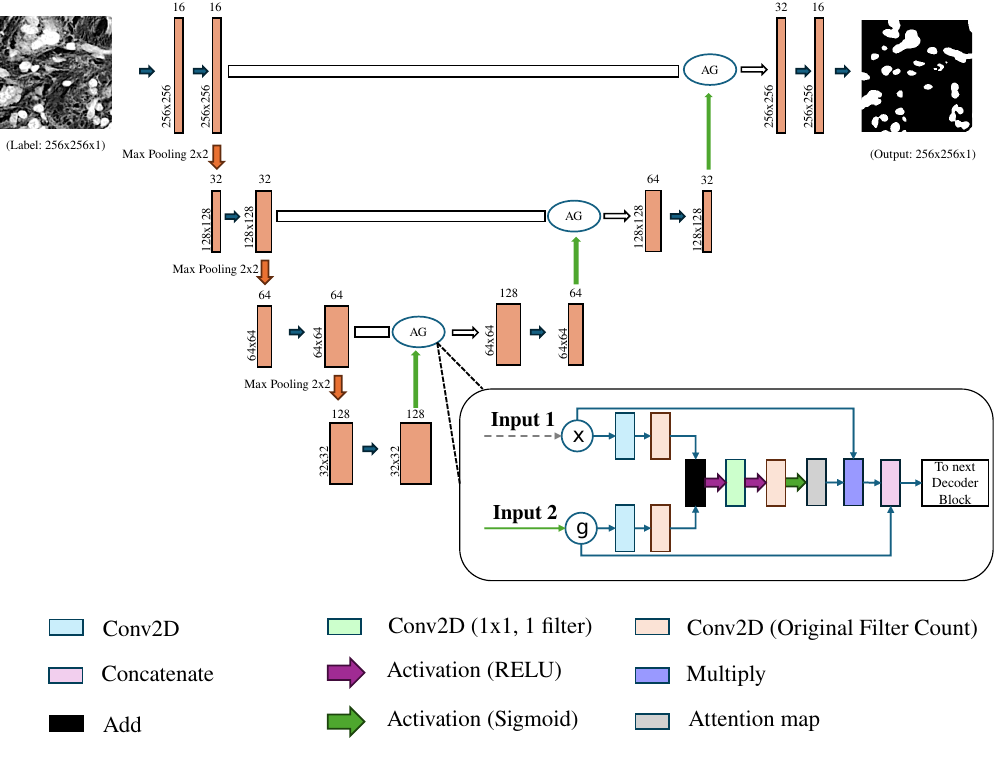}
\caption{Attention U-Net Architecture with Attention Gate.}
\label{fig4}
\end{figure}

\subsubsection{Training and Implementation}
\label{subsubsec:training}

The model was implemented in the TensorFlow framework, version 2.10, using the Keras API. It was then compiled using the Adam optimizer. A custom weighted combined loss, which is the sum of Binary Cross-Entropy (BCE) and Dice Loss \cite{b22,b18}, was used to tackle class imbalance effectively; the Dice Coefficient acted as the main metric of performance during training. The maximum number of epochs was set to 100, while a batch size of 8 was used. To make this model robust and avoid overfitting, several callbacks were used: ModelCheckpoint, which saved the best model based on the validation loss; EarlyStopping, which stopped the training after 5 epochs with no improvement in the metric of interest; and ReduceLROnPlateau reduced the learning rate by a factor of 0.1 if the validation loss reached a plateau after 3 epochs.

\subsubsection{Inference and Evaluation Metrics}
\label{subsubsec:inference}

A full-image inference pipeline was developed to obtain segmentation masks for 1000x1000 pixel test images to evaluate how well the model generalizes. This started with taking a whole test image through precisely the same pre-processing steps taken for the training data. Each preprocessed image was then systematically divided into a grid of overlapping 256x256 patches. Each patch was passed independently through the learned Attention U-Net model for a prediction. These predicted patches were stitched together to reconstruct a full-resolution 1000x1000 mask. In order to make sure that results are smooth, seamless, and with as few edge artifacts as possible, a weighted averaging scheme was applied across overlapping regions.

\begin{figure}[!htbp]
\centering
\includegraphics[width=0.45\textwidth,height=0.42\textheight,keepaspectratio]{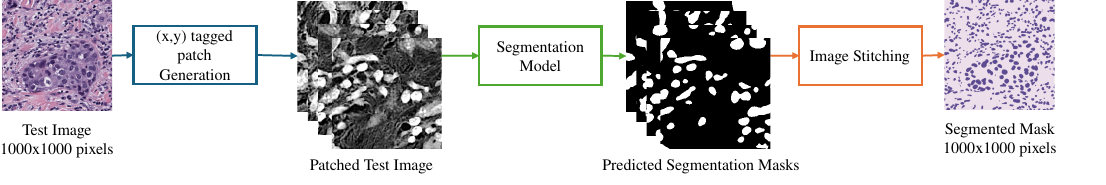}
\caption{Image evaluation pipeline}
\label{fig5}
\end{figure}

The quantitative performance of the final stitched mask was measured against the corresponding ground truth using the IoU metric, which is also known as the Jaccard index \cite{b19}. IoU is a standard metric for the segmentation tasks that quantifies the overlap between the predicted and true masks, calculated as:

\begin{equation}
\mathit{IoU} = \frac{|A \cap B|}{|A \cup B|} = \frac{\mathit{TP}}{\mathit{TP} + \mathit{FP} + \mathit{FN}}
\label{eq:iou}
\end{equation}

\noindent where TP, FP, and FN represent the number of true positive, false positive, and false negative pixels, respectively.

\section{Results}
\label{sec:results}

\subsection{Experimental Analysis}
\label{subsec:experimental}

\begin{figure}[!htbp]
\centering
\includegraphics[width=0.45\textwidth,height=0.42\textheight,keepaspectratio]{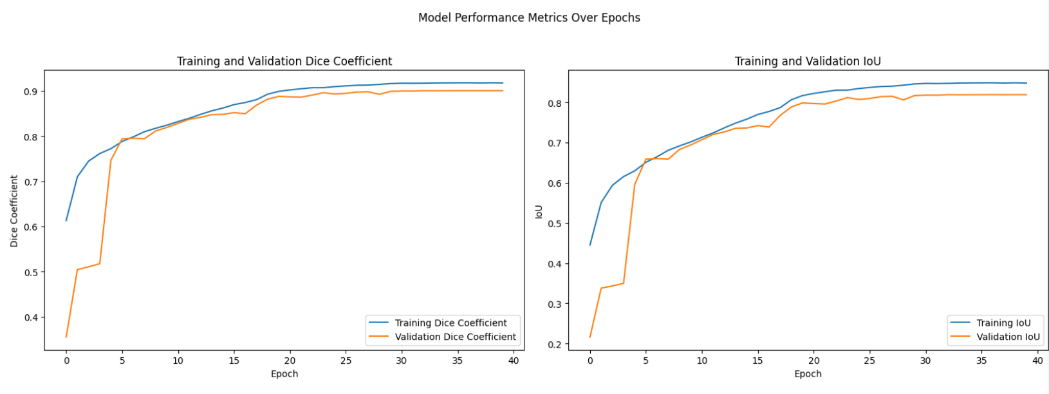}
\caption{Training metrics over epochs}
\label{fig6}
\end{figure}

The Attention U-Net model \cite{b20} was trained and evaluated to see its performance on the MoNuSeg 2018 dataset. The model was compiled with the Adam optimizer and a custom weighted combined loss Binary Cross-Entropy + Dice Loss, addressing class imbalance issues. Consequently, analysis of the training history showed the model steadily improved rapidly in its initial phase, converging around Epoch 5. The best validation performance of the model was obtained at Epoch 37, where the validation IoU of 0.8190 and Dice Coefficient of 0.9004 were recorded and saved as the final model checkpoint. This training was gracefully managed by callbacks: ReduceLROnPlateau prevented early stagnation, while EarlyStopping stopped it during Epoch 40, confirming that the model's capacity had been utilized without severe overfitting. The learning curves for training and validation are presented in Fig.~\ref{fig6}.

For a proper quantification of the model's generalization capability on completely unseen data, the entire patch-prediction and stitching pipeline was conducted on dedicated test images. The stitched full 1000x1000 predictions yielded a final IoU of 0.8745 and a Dice Coefficient of 0.9323. Performance metrics for the proposed method are shown in Table~\ref{tab1}. A comparison of our model's performance against other state-of-the-art methods on the MoNuSeg dataset is given in Table~\ref{tab2}.

\begin{table}[!htbp]
\centering
\caption{Performance metrics for the proposed method}
\label{tab1}
\begin{tabular}{l c c}
\toprule
\textbf{Metric} & \textbf{Mean} & \textbf{Variance} \\
\midrule
IoU              & 0.8745 & 0.0024 \\
Dice Coefficient & 0.9323 & 0.0009 \\
Accuracy         & 0.9721 & 0.0002 \\
Precision        & 0.9067 & 0.0017 \\
Sensitivity      & 0.9602 & 0.0005 \\
Specificity      & 0.9733 & 0.0002 \\
VOI              & 0.3352 & 0.0151 \\
PRI              & 0.8730 & 0.0020 \\
\bottomrule
\end{tabular}
\end{table}

\begin{table}[!htbp]
\centering
\caption{Comparative Analysis of Segmentation Performance on the MoNuSeg 2018 Test Set.}
\label{tab2}
\resizebox{\columnwidth}{!}{%
\begin{tabular}{l c c c}
\toprule
\textbf{Model} & \textbf{Dice Coefficient} & \textbf{Jaccard Index} & \textbf{Reference} \\
\midrule
Baseline U-Net                 & 0.8892 & 0.8031 & \cite{b12} \\
Mask R-CNN                     & 0.8876 & 0.8005 & \cite{b12} \\
U-Net                          & 0.7907 & 0.6826 &            \\
CUMedVision (Challenge Winner) & 0.9079 & 0.8354 & \cite{b12} \\
BiU-net                        & 0.8027 & 0.6722 & \cite{b37} \\
Proposed Attention U-Net       & 0.9323 & 0.8745 & This study \\
\bottomrule
\end{tabular}}
\end{table}

\section{Discussion}
\label{sec:discussion}

\begin{figure}[!htbp]
\centering
\includegraphics[width=0.45\textwidth,height=0.42\textheight,keepaspectratio]{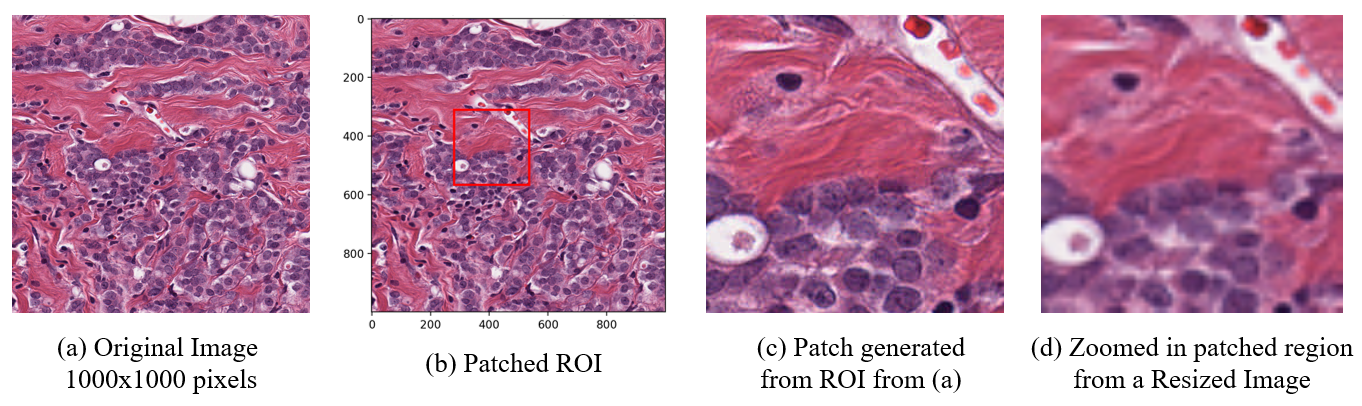}
\caption{Visual representation of information loss in histopathological images during resizing.}
\label{fig7}
\end{figure}

The qualitative and quantitative results obtained validate the effectiveness of the pre-processing pipeline proposed in this study. The performance of the model, hence, can be attributed to methodological decisions. Firstly, the study proposed to build a pipeline for high-resolution patch-based strategy. This was incorporated to preserve minute details of the nuclear boundaries which were lost while downsampling the 1000x1000px images as shown in Fig.~\ref{fig7}, and to serve as a powerful data augmentation step for the 24 images from the training set to 2,400 patches.

The study used stain deconvolution \cite{b23} to isolate the Hematoxylin channel, furthermore Contrast Limited Adaptive Histogram Equalization (CLAHE) \cite{b24} was applied to the images to sharpen the boundary between nucleus and the cytoplasm. Finally, the encapsulated process of generating patches and stitching images as described previously in Fig.~\ref{fig5} preserve fine-grained details.

When generating patches, some patches may have dense cluster of nuclei, while some have sparse nuclei in the background, a standard U-Net's skip connections \cite{b21} might propagate redundant features \cite{b25,b26,b27}, hence, we hypothesized, that attention gates would allow the model to specifically focus on the features enhanced by the preprocessing pipeline incorporated and suppress irrelevant information. By achieving an IoU score that outperformed the baseline U-Net as described in Table~\ref{tab1}, and is highly competitive with the top performing model, the lighter model proposed proved its effectiveness. This demonstrated that the attention mechanism incorporated into the U-Net model allowed for more efficient and focused learning process without need for a complex architecture.

\begin{figure}[!htbp]
\centering
\includegraphics[width=0.5\textwidth,height=0.42\textheight,keepaspectratio]{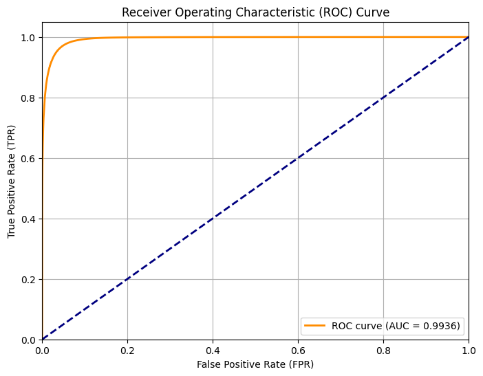}
\caption{Receiver Operating Characteristic (ROC) curve for the Attention U-Net model}
\label{fig8}
\end{figure}

The ROC (Receiver Operating Characteristic) curve in Fig.~\ref{fig8} shows an AUC (Area Under the Curve) value of 0.9936. This high value indicates a very good pixel-level classification. The model is effective in distinguishing between different classes across all threshold values, as demonstrated by the sharp bend at top left corner in the curve. This strong pixel-level accuracy is the foundation for good segmentation performance, as observed by our IoU and Dice Coefficient results.

The use of a combined Binary Cross-Entropy (BCE) \cite{b22} and Dice Loss (DL) \cite{b18} in this study was justified by the inherent class imbalance of the task, where background pixels far outnumbered the nuclei pixels. This custom loss was chosen to ensure that the model performed well on both pixel-level accuracy (BCE) and global segmentation overlap (Dice). The stable learning curves shown in Fig.~\ref{fig6} demonstrate that the model converged to an optimal solution while mitigating the inherent issue of class imbalance.

\section{Conclusion}
\label{sec:conclusion}

This study implemented an Attention U-Net model, that was designed to be lightweight. This model was used to segment nucleus from histopathology images provided by the MoNuSeg 2018 dataset; the model thus trained achieved a competitive IoU score of 0.8265, on the unseen test set images. The methodology consisted of integrating an advanced preprocessing pipeline to extract H and E stain channels from an image and then perform patching to increase the image count. The results validate the approach used in this study as a scalable foundation for other precise qualitative analysis of digital pathologies. Future work should focus on enhancing the current pipeline with incorporating algorithms to find the optimal number of patches to generate for each image \cite{b28,b29,b30}. The preprocessing involved extracting the hematoxylin channel since U-Net models have been shown to struggle with segmentation where a single class's color can vary \cite{b31,b32}, but further color normalization techniques can be explored to enhance image quality \cite{b33,b34,b35,b36}. Focuses on enhancing the generalization through cross-dataset validation and instance segmentation will enable the extraction of more detailed statistics necessary for advanced clinical applications.

\section*{Authors' Contribution}

A.G. and D.G. conceptualized the study, designed the experimental methodology, implemented the proposed architecture and training pipeline, performed experiments, analyzed the results and drafted the initial manuscript.

T.K.M. supervised the research process, provided conceptual guidance, refined the methodology, contributed to interpretation of findings, reviewed and edited the manuscript, and approved the final version of submission.

\bibliographystyle{elsarticle-num}
\begingroup
\footnotesize
\bibliography{references}

@article{b1,
  author  = {Basu, A. and Senapati, P. and Deb, M. and Rai, R. and Dhal, K.},
  title   = {A survey on recent trends in deep learning for nucleus segmentation from histopathology images},
  journal = {Evolving Systems},
  volume  = {14},
  pages   = {1--46},
  year    = {2023},
  doi     = {10.1007/s12530-023-09491-3}
}

@article{b2,
  author  = {Kumar, N. and Verma, R. and Sharma, S. and Bhargava, S. and Vahadane, A. and Sethi, A.},
  title   = {A Dataset and a Technique for Generalized Nuclear Segmentation for Computational Pathology},
  journal = {IEEE Transactions on Medical Imaging},
  volume  = {36},
  number  = {7},
  pages   = {1550--1560},
  year    = {2017},
  doi     = {10.1109/tmi.2017.2677499}
}

@article{b3,
  author  = {Siddique, N. and Paheding, S. and Elkin, C. and Devabhaktuni, V.},
  title   = {{U-Net} and Its Variants for Medical Image Segmentation: A Review of Theory and Applications},
  journal = {IEEE Access},
  volume  = {9},
  pages   = {82031--82057},
  year    = {2021},
  doi     = {10.1109/access.2021.3086020}
}

@article{b4,
  author  = {Wu, J. and Tao, H. and Xiao, K. and Chu, J. and Leng, L.},
  title   = {Multi-stream feature aggregation network with multi-scale supervision for single image dehazing},
  journal = {Engineering Applications of Artificial Intelligence},
  volume  = {139},
  pages   = {109486},
  year    = {2025},
  doi     = {10.1016/j.engappai.2024.109486}
}

@article{b5,
  author  = {Wang, J. and Ruhaiyem, N. and Fu, P.},
  title   = {A Comprehensive Review of {U-Net} and Its Variants: Advances and Applications in Medical Image Segmentation},
  journal = {arXiv preprint arXiv:2502.06895},
  year    = {2025},
  doi     = {10.48550/arxiv.2502.06895}
}

@article{b6,
  author  = {Pan, P. and Zhang, C. and Sun, J. and Guo, L.},
  title   = {Multi-scale conv-attention {U-Net} for medical image segmentation},
  journal = {Scientific Reports},
  volume  = {15},
  pages   = {96101},
  year    = {2025},
  doi     = {10.1038/s41598-025-96101-8}
}

@article{b7,
  author  = {Cao, Y. and Cheng, Y.},
  title   = {{SACU-Net}: Shape-Aware {U-Net} for Biomedical Image Segmentation With Attention Mechanism and Context Extraction},
  journal = {IEEE Access},
  volume  = {13},
  pages   = {5719--5730},
  year    = {2025},
  doi     = {10.1109/access.2025.3526602}
}

@article{b8,
  author  = {Zhang, F. and Fan, Z.},
  title   = {{CS U-NET}: A Medical Image Segmentation Method Integrating Spatial and Contextual Attention Mechanisms Based on {U-NET}},
  journal = {International Journal of Imaging Systems and Technology},
  volume  = {35},
  number  = {1},
  pages   = {e70072},
  year    = {2025},
  doi     = {10.1002/ima.70072}
}

@article{b9,
  author  = {Kamiri, J. and Wambugu, G. and Oirere, A.},
  title   = {An Empirical Analysis of Encoder-Decoder ({U-Net}) Variants for Medical Image Segmentation},
  journal = {International Journal of Scientific Research in Computer Science and Engineering},
  volume  = {13},
  number  = {1},
  pages   = {6--10},
  year    = {2025},
  doi     = {10.26438/ijsrcse.v13i1.610}
}

@article{b10,
  author  = {Abdulfatah, A. and Sheng, Z. and Tenawerk, Y.},
  title   = {{U-Net}-Based Medical Image Segmentation: A Comprehensive Analysis and Performance Review},
  journal = {Journal of Electronic Research and Application},
  volume  = {9},
  number  = {1},
  pages   = {9450},
  year    = {2025},
  doi     = {10.26689/jera.v9i1.9450}
}

@article{b11,
  author  = {Chen, J. and Mei, J. and Li, X. and Lu, Y. and Yu, Q. and Wei, Q. and Luo, X. and Xie, Y. and Adeli, E. and Wang, Y. and Lungren, M. and Zhang, S. and Xing, L. and Lu, L. and Yuille, A. and Zhou, Y.},
  title   = {{TransUNet}: Rethinking the {U-Net} architecture design for medical image segmentation through the lens of transformers},
  journal = {Medical Image Analysis},
  volume  = {97},
  pages   = {103280},
  year    = {2024},
  doi     = {10.1016/j.media.2024.103280}
}

@article{b12,
  author  = {Kumar, N. and Verma, R. and Anand, D. and Zhou, Y. and others},
  title   = {A Multi-Organ Nucleus Segmentation Challenge},
  journal = {IEEE Transactions on Medical Imaging},
  volume  = {39},
  number  = {5},
  pages   = {1380--1391},
  year    = {2020},
  doi     = {10.1109/tmi.2019.2947628}
}

@article{b13,
  author  = {Verma, R. and Kumar, N. and Patil, A. and Kurian, N. and Rane, S. and others},
  title   = {{MoNuSAC2020}: A Multi-Organ Nuclei Segmentation and Classification Challenge},
  journal = {IEEE Transactions on Medical Imaging},
  volume  = {40},
  number  = {12},
  pages   = {3413--3423},
  year    = {2021},
  doi     = {10.1109/tmi.2021.3085712}
}

@article{b16,
  author  = {Yao, K. and Huang, K. and Sun, J. and Jing, L. and Huang, D. and Jude, C.},
  title   = {{Scaffold-A549}: A Benchmark {3D} Fluorescence Image Dataset for Unsupervised Nuclei Segmentation},
  journal = {Cognitive Computation},
  volume  = {13},
  number  = {6},
  pages   = {1603--1608},
  year    = {2021},
  doi     = {10.1007/s12559-021-09944-4}
}

@article{b18,
  author  = {Dice, L. R.},
  title   = {Measures of the Amount of Ecologic Association Between Species},
  journal = {Ecology},
  volume  = {26},
  number  = {3},
  pages   = {297--302},
  year    = {1945},
  doi     = {10.2307/1932409}
}

@article{b19,
  author  = {Jaccard, P.},
  title   = {The distribution of the flora in the alpine zone},
  journal = {New Phytologist},
  volume  = {11},
  number  = {2},
  pages   = {37--50},
  year    = {1912},
  doi     = {10.1111/j.1469-8137.1912.tb05611.x}
}

@article{b20,
  author  = {Oktay, O. and Schlemper, J. and Folgoc, L. and Lee, M. and Heinrich, M. and Misawa, K. and Mori, K. and McDonagh, S. and Hammerla, N. and Kainz, B. and Glocker, B. and Rueckert, D.},
  title   = {Attention {U-Net}: Learning Where to Look for the Pancreas},
  journal = {arXiv preprint arXiv:1804.03999},
  year    = {2018}
}

@inproceedings{b21,
  author    = {Ronneberger, O. and Fischer, P. and Brox, T.},
  title     = {{U-Net}: Convolutional Networks for Biomedical Image Segmentation},
  booktitle = {Medical Image Computing and Computer-Assisted Intervention -- MICCAI 2015},
  publisher = {Springer},
  address   = {Cham},
  pages     = {234--241},
  year      = {2015},
  doi       = {10.1007/978-3-319-24574-4_28}
}

@inproceedings{b22,
  author    = {Liu, L. and Qi, H.},
  title     = {Learning Effective Binary Descriptors via Cross Entropy},
  booktitle = {2017 IEEE Winter Conference on Applications of Computer Vision (WACV)},
  pages     = {1251--1258},
  year      = {2017},
  doi       = {10.1109/wacv.2017.144}
}

@article{b23,
  author  = {Landini, G. and Martinelli, G. and Piccinini, F.},
  title   = {Colour deconvolution: stain unmixing in histological imaging},
  journal = {Bioinformatics},
  volume  = {37},
  number  = {14},
  pages   = {2050--2057},
  year    = {2021},
  doi     = {10.1093/bioinformatics/btaa847}
}

@inproceedings{b24,
  author    = {Pizer, S. M. and Johnston, R. E. and Ericksen, J. P. and Yankaskas, B. C. and Muller, K. E.},
  title     = {Contrast-limited adaptive histogram equalization: speed and effectiveness},
  booktitle = {[1990] Proceedings of the First Conference on Visualization in Biomedical Computing},
  pages     = {337--345},
  year      = {1990},
  doi       = {10.1109/vbc.1990.109340}
}

@article{b25,
  author  = {Zhang, J. and Jiang, Z. and Dong, J. and Hou, Y. and Liu, B.},
  title   = {Attention Gate {ResU-Net} for Automatic {MRI} Brain Tumor Segmentation},
  journal = {IEEE Access},
  volume  = {8},
  pages   = {58533--58545},
  year    = {2020},
  doi     = {10.1109/access.2020.2983075}
}

@article{b26,
  author  = {Xu, Z. and Tian, B. and Liu, S. and Wang, X. and Yuan, D. and Gu, J. and Chen, J. and Lukasiewicz, T. and Leung, V. C. M.},
  title   = {Collaborative Attention Guided Multi-Scale Feature Fusion Network for Medical Image Segmentation},
  journal = {IEEE Transactions on Network Science and Engineering},
  volume  = {11},
  number  = {2},
  pages   = {1857--1871},
  year    = {2024},
  doi     = {10.1109/tnse.2023.3332810}
}

@article{b27,
  author  = {Lian, S. and Luo, Z. and Zhong, Z. and Lin, X. and Su, S. and Li, S.},
  title   = {Attention guided {U-Net} for accurate iris segmentation},
  journal = {Journal of Visual Communication and Image Representation},
  volume  = {56},
  pages   = {296--304},
  year    = {2018},
  doi     = {10.1016/j.jvcir.2018.10.001}
}

@article{b28,
  author  = {Tang, Y. and Gao, R. and Lee, H. and Han, S. and Chen, Y. and Gao, D. and Nath, V. and Bermudez, C. and Savona, M. R. and Abramson, R. G. and Bao, S. and Lyu, I. and Huo, Y. and Landman, B. A.},
  title   = {High-resolution {3D} abdominal segmentation with random patch network fusion},
  journal = {Medical Image Analysis},
  volume  = {69},
  pages   = {101894},
  year    = {2021},
  doi     = {10.1016/j.media.2020.101894}
}

@article{b29,
  author  = {Jenkinson, E. and Arandjelovi{\'c}, O.},
  title   = {Whole Slide Image Understanding in Pathology: What Is the Salient Scale of Analysis?},
  journal = {BioMedInformatics},
  volume  = {4},
  number  = {1},
  pages   = {490--509},
  year    = {2024},
  doi     = {10.3390/biomedinformatics4010028}
}

@article{b30,
  author  = {Lou, W. and Li, H. and Li, G. and Han, X. and Wan, X.},
  title   = {Which Pixel to Annotate: A Label-Efficient Nuclei Segmentation Framework},
  journal = {IEEE Transactions on Medical Imaging},
  volume  = {42},
  number  = {4},
  pages   = {947--958},
  year    = {2023},
  doi     = {10.1109/tmi.2022.3221666}
}

@inproceedings{b31,
  author    = {Pan, F. and Geng, L. and Zhang, N. and Chen, Z.},
  title     = {Improved {U-Net} with Gray Channel Attention for Image Segmentation},
  booktitle = {2024 4th International Conference on Intelligent Communications and Computing (ICICC)},
  pages     = {70--73},
  year      = {2024},
  doi       = {10.1109/icicc63565.2024.10780506}
}

@article{b32,
  author  = {Azad, R. and Aghdam, E. K. and Rauland, A. and Jia, Y. and Avval, A. H. and Bozorgpour, A. and Karimijafarbigloo, S. and Cohen, J. P. and Adeli, E. and Merhof, D.},
  title   = {Medical Image Segmentation Review: The Success of {U-Net}},
  journal = {IEEE Transactions on Pattern Analysis and Machine Intelligence},
  volume  = {46},
  number  = {12},
  pages   = {10076--10095},
  year    = {2024},
  doi     = {10.1109/tpami.2024.3435571}
}

@inproceedings{b33,
  author    = {Yildirim, Z. and Han{\c{c}}er, E. and Samet, R. and Mali, M. and Nemati, N.},
  title     = {Effect of Color Normalization on Nuclei Segmentation Problem in {H\&E} Stained Histopathology Images},
  booktitle = {2022 30th Signal Processing and Communications Applications Conference (SIU)},
  pages     = {1--4},
  year      = {2022},
  doi       = {10.1109/siu55565.2022.9864814}
}

@article{b34,
  author  = {Ding, H. and Cen, Q. and Si, X. and Pan, Z. and Chen, X.},
  title   = {Automatic glottis segmentation for laryngeal endoscopic images based on {U-Net}},
  journal = {Biomedical Signal Processing and Control},
  volume  = {71},
  pages   = {103116},
  year    = {2022},
  doi     = {10.1016/j.bspc.2021.103116}
}

@article{b35,
  author  = {Javvadi, S.},
  title   = {Evaluating the Impact of Color Normalization on Kidney Image Segmentation},
  journal = {International Journal on Cybernetics \& Informatics},
  volume  = {12},
  number  = {5},
  pages   = {95--104},
  year    = {2023},
  doi     = {10.5121/ijci.2023.120509}
}

@article{b36,
  author  = {Boschman, J. and Farahani, H. and Darbandsari, A. and Ahmadvand, P. and Van Spankeren, A. and Farnell, D. J. and Levine, A. B. and Naso, J. R. and Churg, A. and Jones, S. J. and Yip, S. and Koebel, M. and Huntsman, D. G. and Gilks, C. B. and Bashashati, A.},
  title   = {The utility of color normalization for {AI}-based diagnosis of hematoxylin and eosin-stained pathology images},
  journal = {The Journal of Pathology},
  volume  = {256},
  number  = {1},
  pages   = {15--24},
  year    = {2022},
  doi     = {10.1002/path.5797}
}

@article{b37,
  author  = {Huang, Z. and Zhao, Y. and Yu, Z. and Qin, P. and Han, X. and Wang, M. and Liu, M. and Gregersen, H.},
  title   = {{BiU-net}: A dual-branch structure based on two-stage fusion strategy for biomedical image segmentation},
  journal = {Computer Methods and Programs in Biomedicine},
  volume  = {252},
  pages   = {108235},
  year    = {2024},
  doi     = {10.1016/j.cmpb.2024.108235}
}
\endgroup

\end{document}